\documentclass[twocolumn]{article}

\usepackage[final]{graphicx} 
\usepackage[utf8]{inputenc} 
\usepackage[T1]{fontenc}    
\usepackage{geometry}       
\usepackage{authblk} 
\usepackage{cuted} 

\author[1]{\textbf{Anis Ur Rahman}}
\author[2]{\textbf{Mete Ahishali}}
\author[3]{\textbf{Einari Heinaro}}
\author[2,4]{\textbf{Samuli Junttila}}
\affil[1]{CSC -- IT Center for Science Ltd., Espoo, Finland}
\affil[2]{Department of Forest Sciences, University of Helsinki}
\affil[3]{KOKO Forest Ltd.}
\affil[4]{School of Forest Sciences, University of Eastern Finland}
\affil[ ]{\texttt{anis.rahman@csc.fi, mete.ahishali@helsinki.fi, einari.heinaro@kokoforest.com, samuli.junttila@\{helsinki,uef\}.fi}}

\usepackage{amsmath, amssymb, amsthm} 
\usepackage{booktabs}                 
\usepackage{microtype}                
\usepackage{url}                      
\usepackage{hyperref}                 
\hypersetup{colorlinks=true, linkcolor=blue, urlcolor=cyan, citecolor=green}

\usepackage{natbib} 
\setcitestyle{numbers,square}

\title{\textbf{Cross-Domain Dead Tree Detection via Knowledge Distillation in Aerial Imagery}}

\date{} 

\newcommand{\mypar}[1]{\noindent\textbf{#1}}

\usepackage{makecell}

\usepackage{duckuments}
\usepackage{comment}
\usepackage{xspace}

\usepackage{lineno}

\usepackage{amsmath}
\usepackage{amsfonts}

\usepackage{booktabs}
\usepackage{tabularx}
\usepackage{adjustbox}
\usepackage{multicol}
\usepackage{multirow}
\usepackage{caption}

\usepackage{subcaption}

\newcommand{\oursone}{\textit{TreeMort-1T-UNet}\xspace}
\newcommand{\oursthree}{\textit{TreeMort-3T-UNet}\xspace}

\usepackage{pifont}

\usepackage[table]{xcolor}
\definecolor{lncolor}{HTML}{FEE4C4}
\definecolor{baseline}{HTML}{E0E4E8}
\definecolor{ppink}{rgb}{0.98, 0.575, 0.89}

\begin{document}

\twocolumn[
  \begin{center}
    \maketitle
    \begin{abstract}
    Detecting dead trees in aerial imagery is vital for assessing forest health, especially as tree mortality increases globally due to climate change, but domain variability and scarce labeled data often limit model generalization. This study advances the \oursone (Tree Mortality 1-Task U-Net) model, initially trained on Finnish aerial imagery (source domain), by applying knowledge distillation (KD) to adapt it to various target domains, including Polish, German, and Estonian datasets representing diverse forest types. We assess four KD variants: Basic, Self, Feature-level, and Ensemble, against a fine-tuning baseline, using Mean Tree IoU, Instance F1-score, Instance Precision, and Mean Centroid Error as key metrics, alongside representational analyses (e.g., cosine similarity, CKA, SSIM, t-SNE, and linear probing) for domain invariance. \textit{Feature-level KD} outperforms others, yielding a Mean Tree IoU of 0.106, Instance F1-score of 0.63, Instance Precision of 0.55, and Mean Centroid Error of 3.039 on the Polish dataset, with robust precision across other target domains (e.g., 0.15 on Finnish, 0.67 on Polish, 0.60 on German, 0.59 on Estonian). It excels in low-data scenarios with fewer false positives and shows superior representational invariance (e.g., higher deep-layer CKA/SSIM, better domain mixing in t-SNE, and linear probing AUC of 0.95), making it ideal for precision-critical forestry applications. Additional ablation studies confirm that key components like feature alignment enhance its performance balance across metrics. Our findings demonstrate KD's potential to enhance transfer learning in remote sensing, offering a scalable, domain-robust tool for ecological monitoring and sustainable forest management.
    \end{abstract}
    \vspace{0.3in}
  \end{center}
]




\section{Introduction}
\label{sec:intro}

Dead trees are critical indicators of forest health, biodiversity, and wildfire risk, with global tree mortality increasing due to climate change-driven factors such as intensified droughts, heat extremes, insect outbreaks, and wildfires, leading to widespread forest die-off and urgent needs for enhanced monitoring~\citep{junttila2024significant,cheng2024scattered,international2025towards}. Their detection through remote sensing is therefore essential for environmental monitoring and effective forestry management, as field surveys are often infeasible due to limited accessibility, high costs, and the scattered occurrence of dead trees across vast areas, while remote sensing offers efficiency, scalability, and comprehensive coverage for mapping individual dead trees. High-resolution aerial imagery enables the precise identification of individual dead trees across large forest areas, revealing mortality patterns increasingly linked to climate change~\citep{ahishali2025ada}. However, this task presents significant challenges. Dense canopy cover and spectral similarities between living and dead vegetation often hinder accurate segmentation~\citep{briechle2021silvi}. Moreover, \textbf{domain variability} poses a substantial obstacle, as models trained on one forest region frequently fail to generalize to others due to differences in tree species, terrain, and imaging conditions~\citep{angarano2024domain}. For instance, dead conifers in Finland's sparse boreal forests are easily distinguishable against open canopies, whereas decaying broadleaves in Poland's dense temperate forests are often obscured by surrounding foliage and variable lighting. These challenges are compounded by \textbf{data scarcity}, as annotated datasets are limited by the high cost of expert labeling and the infrequent occurrence of dead trees in certain regions~\citep{kellenberger2018detecting, liu2019knowledge}.

Traditional transfer learning methods, such as fine-tuning pre-trained models, often prove inadequate in this context. Fine-tuning requires labeled data from the target domain, which is frequently unavailable, and it struggles with significant distribution shifts, leading to overfitting or the loss of generalizable features~\citep{tuia2016domain}. Recent domain adaptation techniques in remote sensing, such as image-to-image translation, aim to align data across different sites~\citep{wang2019domain, ahishali2025ada}. Although these methods have shown promise for tasks like cross-site tree mortality mapping~\citep{chen2022object, dalponte2019individual}, they often involve complex training processes and fail to fully leverage the knowledge embedded in segmentation models.

To overcome these limitations, we propose \textbf{knowledge distillation (KD)} as an effective strategy for improving cross-domain generalization in dead tree segmentation. KD enables a compact student model to learn from a robust teacher model, absorbing both class labels and subtle insights from the teacher's outputs~\citep{hinton2015distilling}. Through this process, the student model learns to capture domain-invariant features, such as textural signs of decay, while adapting to region-specific differences. Unlike fine-tuning, KD uses a teacher model trained on a well-annotated source domain (e.g., Finnish aerial imagery) to guide the student's learning on an unseen target domain (e.g., Polish imagery), reducing the need for extensive retraining. This approach not only addresses domain variability but also mitigates the challenges posed by scarce labeled data, making KD a scalable solution for ecological monitoring.

In this study, we evaluate the effectiveness of KD using datasets from Finland and Poland, which represent boreal and temperate forest ecosystems, respectively. Our findings demonstrate KD's capacity to bridge domain gaps and reduce the dependence on site-specific annotations, thereby advancing the application of remote sensing in forestry. Specifically, our contributions are as follows:

\begin{enumerate}
\item \textbf{Pioneering KD Application:} This research introduces KD as a novel strategy for dead tree detection, tackling domain variability and data scarcity in remote sensing.
\item \textbf{Quantifying Domain Variability}: We examine how domain shifts impact dead tree detection, underscoring the need for robust generalization techniques.
\item \textbf{Comparative KD Evaluation}: We demonstrate KD's advantages over traditional transfer learning across diverse aerial imagery domains.
\item \textbf{Practical Implications}: We develop efficient, distilled models to support scalable forest health monitoring and wildfire risk assessment.
\end{enumerate}

The paper is structured as follows: Section~\ref{sec:related} reviews related work, Section~\ref{sec:method} outlines our methodology, Section~\ref{sec:results} presents experimental findings, Section~\ref{sec:discussion} discusses broader implications, and Section~\ref{sec:conclusion} provides concluding remarks.

\section{Related Work}
\label{sec:related}

Transfer learning and knowledge distillation have emerged as pivotal techniques in addressing the challenges of data scarcity and computational constraints in remote sensing, particularly within the domain of ecological monitoring. This section provides a comprehensive overview of the existing literature, highlighting the evolution of these methods and their application to tasks analogous to dead tree detection in aerial imagery.

\subsection{Transfer Learning in Remote Sensing}

Transfer learning, which leverages pre-trained models to adapt to new tasks or domains, has become a cornerstone in remote sensing due to the limited availability of labeled datasets. Models pre-trained on large, general-purpose datasets such as \textit{ImageNet}~\citep{deng2009imagenet} are frequently fine-tuned for specific remote sensing tasks, including land cover classification~\citep{xie2016transfer}, object detection~\citep{chen2022object}, and semantic segmentation~\citep{zhang2016deep}. For instance, Xie et al.~\citep{xie2016transfer} demonstrated that fine-tuning a \textit{CNN} pre-trained on \textit{ImageNet} significantly improves classification accuracy on satellite imagery, even with small training sets. However, the efficacy of such approaches diminishes when confronted with pronounced domain shifts, as is common in forestry applications where spectral and structural variability across regions can be substantial~\citep{tuia2016domain}.

In the context of forestry, transfer learning has been applied to tasks such as tree species classification~\citep{kellenberger2018detecting} and estimation of forest structural variables~\citep{fassnacht2016review}. \cite{kellenberger2018detecting} utilized a fine-tuned \textit{ResNet} model to classify tree species in UAV imagery, achieving notable accuracy improvements over training from scratch. Nonetheless, these studies often assume a degree of domain similarity that may not hold for dead tree detection, where the visual signatures of decay are subtle and highly variable across ecosystems.

\subsection{Knowledge Distillation: Principles and Applications}

Knowledge distillation~\citep{hinton2015distilling} extends the paradigm of transfer learning by enabling the transfer of learned representations from a larger, pre-trained ``teacher" model to a smaller, more efficient ``student" model. This technique not only facilitates model compression but also enhances generalization by leveraging the teacher's so-called \textit{dark knowledge}, i.e., the implicit relationships encoded in its output probabilities. Distillation has been widely adopted in computer vision for tasks such as image classification~\citep{romero2014fitnets} and object detection~\citep{chen2017learning}, where it has been shown to improve both accuracy and inference speed.

In remote sensing, knowledge distillation has been explored for applications including land cover mapping~\citep{liu2019knowledge} and hyperspectral image analysis~\citep{wang2019domain}. For example, \cite{liu2019knowledge} employed distillation to compress a large \textit{CNN} for land cover classification, achieving comparable accuracy with a fraction of the computational cost. However, the application of distillation to more specialized tasks, such as dead tree detection, remains underexplored. The unique challenges of this task, such as the need to discern fine-grained, domain-specific features, necessitate tailored distillation strategies that go beyond standard output-level supervision.

\subsection{Dead Tree Detection: Challenges and Existing Approaches}

The detection of dead trees in aerial imagery is a particularly demanding task due to several factors:

\begin{itemize}
\item \textbf{Spectral Similarity:} Dead trees often exhibit spectral signatures that closely resemble those of living trees or other vegetation in early decay stages~\citep{meddens2011evaluating}, but in later stages, they may be more similar to bare ground or rocks due to foliage loss, particularly in infrared regions, leading to classification challenges~\citep{zielewska2020detection,liu2021mapping}.
\item \textbf{Occlusion and Clutter:} Dense forest canopies can obscure dead trees, while cluttered backgrounds complicate segmentation~\citep{dalponte2019individual}.
\item \textbf{Domain Variability:} Variations in tree species, terrain, imaging conditions, and decay stages across different regions exacerbate the generalization problem~\citep{fassnacht2014comparison}. For instance, dead trees in early decay resemble living vegetation, while long-standing dead trees with few remaining branches are notably harder to detect due to their sparse appearance.
\end{itemize}

Existing methods for dead tree detection range from traditional image processing techniques, such as thresholding and edge detection~\citep{garrity2013quantifying}, to more advanced machine learning approaches~\citep{polewski2018learning}. However, these methods frequently rely on handcrafted features or require extensive domain-specific tuning, limiting their applicability across diverse forest ecosystems. Recent studies have begun to explore deep learning for this task~\citep{briechle2021silvi}, yet they often overlook the critical issue of domain adaptation, assuming access to large, region-specific labeled datasets.

\subsection{Positioning of the Current Work}

This study addresses the aforementioned gaps by investigating the efficacy of knowledge distillation for dead tree detection in aerial imagery, with a particular emphasis on cross-domain generalization. Unlike prior work that primarily focuses on fine-tuning or standard distillation, we implement and rigorously evaluate four distinct distillation variants: \textit{Basic KD}, \textit{Self-KD}, \textit{Feature-level KD}, and \textit{Ensemble KD}. Each approach is designed to tackle specific aspects of the detection challenge. Notably, our \textit{Feature-level KD} approach incorporates intermediate feature alignment to capture domain-robust representations, a strategy that has not been extensively explored in the context of forestry applications. Furthermore, we benchmark these methods against a fine-tuning baseline, providing a comprehensive analysis of their relative strengths and weaknesses.

Through this work, we aim to not only advance the state of the art in dead tree detection but also to contribute to the broader discourse on transfer learning in remote sensing, offering insights that may be applicable to a range of ecological monitoring tasks.

\section{Methodology}
\label{sec:method}

\begin{figure*}[!ht]
    \centering
    \includegraphics[width=\linewidth]{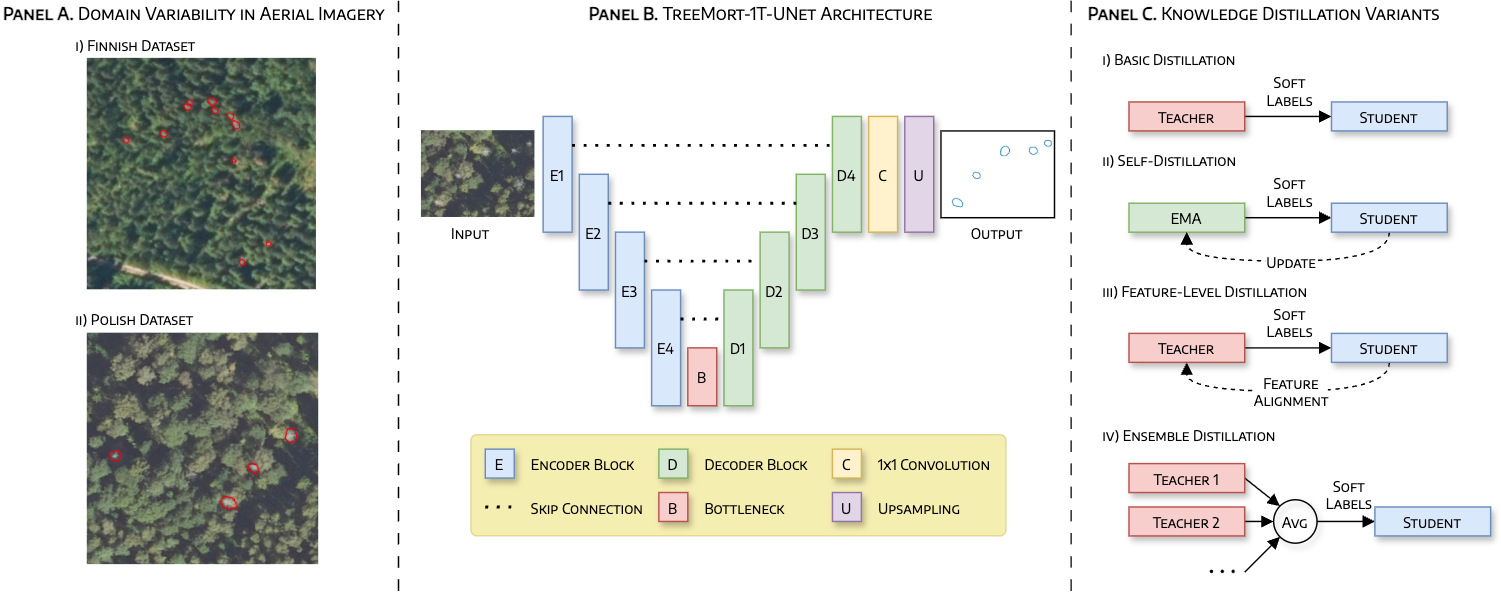}
    \caption{
        Overview of the methodology for cross-domain dead tree detection in aerial imagery, presented in three panels. 
        Panel A illustrates the variability in the dataset, showcasing differences in forest density, tree species, and imaging conditions between Finnish and Polish aerial imagery, which pose significant challenges for domain adaptation in dead tree detection. 
        Panel B depicts the \oursone model architecture, highlighting its \textit{ResNet-34} encoder, self-attention-based decoder, and skip connections, designed for robust feature extraction and segmentation in multi-spectral aerial imagery. 
        Panel C presents the four KD variants: Basic, Self, Feature-level, and Ensemble, illustrating their mechanisms for transferring knowledge from teacher to student models to address domain variability and data scarcity.
    }
    \label{fig:overview}
\end{figure*}

This section outlines the methodology developed for detecting dead trees in aerial imagery, leveraging knowledge distillation to optimize model performance and adaptability across diverse forest ecosystems. The approach encompasses a comprehensive dataset preparation and preprocessing workflow, a carefully designed model architecture, multiple knowledge distillation strategies, a fine-tuning baseline for comparison, and rigorous training and evaluation procedures. Each component is tailored to address the unique challenges of dead tree detection, aiming to deliver models that are both accurate and generalizable. An overview of the methodology, including the dataset variability, model architecture, and KD variants, is illustrated in Figure~\ref{fig:overview}.

\subsection{Dataset and Preprocessing}

The research uses high-resolution aerial imagery from four distinct regions: Finland, Poland, Germany, and Estonia, as summarized in Table~\ref{tbl:dataset}. The Finnish dataset, sourced from the National Land Survey of Finland~\citep{nlsf2024}, includes orthophotos captured between 2011 and 2023 with a true spatial resolution of 0.5 meters per pixel (resampled to 0.25 meters per pixel for this study), covering approximately 100 km$^2$ of diverse forest terrain. This dataset serves as the source domain for training the teacher model and contains approximately 17,500 manually annotated dead tree polygons (each consisting of a single dead tree), labeled by our collaborating forest health experts. The Polish dataset~\citep{gugik_orto}, used as the target domain for adaptation, comprises aerial imagery from Polish forests, which was standardized to the same resolution during preprocessing. Similarly, standing dead trees were manually annotated as polygons, around $9,700$ in total. The German dataset, sourced from Open NRW~\citep{open_nrw_2024}, and the Estonian dataset from the Estonian Land Board~\citep{maaamet_2024}, are used exclusively for evaluation to assess cross-domain generalization; they contain around 4,200 and 1,800 annotated dead trees, respectively, covering smaller areas with higher tree densities. The Estonian orthophotos feature resolutions of 0.20--0.40 m (general) and 0.10--0.16 m (settlements), with annual updates and CIR (color-infrared: NIR, red, green) bands suitable for forestry monitoring.

A multi-channel labeling scheme enhances detection and localization precision. The proposed scheme first employs a binary mask to signify dead tree presence. Then, a Gaussian heatmap is centered on tree centroids for accurate positioning, and a hybrid channel integrating the Signed Distance Transform with boundary details refines edge delineation. Given the large scale of the imagery, a patch-based approach segments images and labels into $256 \times 256$ pixel patches with a 128-pixel stride, ensuring 50\% overlap to maintain contextual integrity. A buffer mask excludes incomplete tree segments from each patch, guaranteeing that only fully represented trees contribute to training. These patches, stored in an HDF5 file, are organized by image and patch index, encapsulating multispectral RGB-NIR image data, label channels, and metadata such as tree counts and geographic coordinates.

Preprocessing normalizes pixel values to a $[0, 1]$ range using histogram truncation, clipping the top and bottom 2\% of values to mitigate outliers. Patches are uniformly sized via center cropping or padding, and an optional image processor adjusts the mean and standard deviation for multi-channel inputs. As data augmentation, we use random flips, rotations, brightness and contrast shifts, multiplicative noise, and gamma correction, which enhances model robustness by simulating varied imaging conditions. To counter class imbalance, where most patches lack dead trees, we have followed a balanced sampling approach ensuring equitable selection of patches with and without dead trees during training.

\begin{table*}[t!]
    \def\arraystretch{1.1}
    \centering
    \footnotesize
    \caption{Overview of datasets and key attributes used for model training and evaluation. Germany and Estonia are evaluation-only datasets.}
    \begin{adjustbox}{max width=\textwidth}
    \begin{tabular}{llcccc}
    \toprule
    ~ & 
    \textsc{\shortstack{Attribute}} & 
    \textsc{\shortstack{Finland}} & 
    \textsc{\shortstack{Poland}} & 
    \textsc{\shortstack{Germany}} & 
    \textsc{\shortstack{Estonia}} \\
    \midrule
    \multicolumn{6}{l}{\texttt{Common Acquisition Parameters}} \\
    & Spectral Bands & \multicolumn{4}{c}{RGB, NIR} \\
    & Patch Size & \multicolumn{4}{c}{$256 \times 256$ px} \\
    & Patch Overlap & \multicolumn{4}{c}{50\%} \\
    \midrule
    \multicolumn{6}{l}{\texttt{Image and Annotation Statistics}} \\
    & Spatial Resolution & 0.25 m & 0.25 m & 0.10 m & 0.25 m \\
    & Image Count & 126 & 166 & 30 & 15 \\
    & Area (km\textsuperscript{2}) & 98.49 & 33.15 & 7.68 & 3.99 \\
    & Acquisition Years & 2011, 2013--17, 2019, 2022--23 & 2021--22 & 2021--22 & 2022--23 \\
    & Dead Trees (annotated) & 22,772 & 10,205 & 4,188 & 2,922 \\
    & Avg. Dead Tree Density (/ha) & 2.31 & 3.08 & 5.46 & 7.33 \\
    \midrule
    \multicolumn{6}{l}{\texttt{Patch-Level Dataset Partitions}} \\
    & Patch Count & 87,458 & 17,648 & 7,264 & 3,128 \\
    & Labeled Trees & 624,966 & 27,087 & 14,616 & 4,915 \\
    & Train / Val / Test Trees & 454k / 127k / 43k & 16.0k / 8.3k / 2.8k & -- & -- \\
    \bottomrule
    \end{tabular}
    \end{adjustbox}
\label{tbl:dataset}
\end{table*}

\subsection{Model Architecture}

Our proposed model, \oursone (Tree Mortality 1-Task U-Net), is an adaptation of the \oursthree  (Tree Mortality 3-Task U-Net) architecture originally introduced in~\citep{rahman2025dual} for multi-task learning with three output channels, reconfigured here for binary dead tree segmentation with a single output mask to streamline computational demands while retaining high accuracy in identifying dead trees from aerial imagery. Built on the U-Net framework, it employs a \textit{ResNet-34} encoder, pre-trained on \textit{ImageNet} and adapted for four-channel RGB-NIR inputs, paired with a decoder featuring self-attention mechanisms to capture long-range dependencies, enhancing segmentation precision, and skip connections to preserve critical spatial information. In our knowledge distillation setup, both the teacher and student models are instances of \oursone, sharing the same architecture to isolate the effects of knowledge distillation on cross-domain adaptation. The teacher model is trained on the Finnish dataset using a hybrid loss function combining binary cross-entropy and Dice loss, optimized with the Adam algorithm~\citep{kingma2014adam}. The student model, initialized with distinct weights, is trained on the Polish dataset, utilizing the teacher's knowledge to adapt to this target domain. Future studies could explore employing a smaller or shallower student model to optimize computational efficiency.

\subsection{Knowledge Distillation Variants}

Knowledge distillation (KD) is a technique designed to transfer knowledge from a pre-trained teacher model to a student model, optimizing performance while addressing challenges such as domain variability and limited labeled data. In this study, we apply four KD variants: Basic, Self, Feature-level, and Ensemble, to adapt the \oursone model from Finnish to Polish aerial imagery for dead tree detection. These variants are tailored to the unique demands of remote sensing, where spectral and structural differences across regions complicate model generalization. Below, we detail each variant's conceptual basis, mathematical formulation, and implementation specifics.

\begin{enumerate}
\item\textbf{Basic KD} trains the student model to balance two objectives: matching ground truth labels and mimicking the teacher's softened predictions. The total loss is a weighted combination of the standard cross-entropy loss ($\mathcal{L}_{\text{CE}}$) and the distillation loss ($\mathcal{L}_{\text{KD}}$):
\begin{align}
\mathcal{L} = \alpha \mathcal{L}_{\text{KD}} + (1 - \alpha) \mathcal{L}_{\text{CE}}.
\end{align}

Here, $\alpha$ increases linearly from 0.3 to 0.7 over training epochs, gradually shifting focus toward the teacher's guidance. The distillation loss employs temperature-scaled probabilities to capture the teacher's nuanced confidence:
\begin{align}
\mathcal{L}_{\text{KD}} = T^2 \cdot \text{BCE} \left( \sigma \left( \frac{\mathbf{z}_s}{T} \right), \sigma \left( \frac{\mathbf{z}_t}{T} \right) \right),
\end{align}
where $\mathbf{z}_s$ and $\mathbf{z}_t$ are the student and teacher logits, respectively, and $\sigma$ is the sigmoid function. The temperature $T$ decreases from 6.0 to 2.0, refining the knowledge transfer as training progresses. A confidence mask ($\sigma(\mathbf{z}_t / T) > 0.3$) ensures the loss focuses on reliable teacher predictions, addressing the uncertainty in aerial imagery where dead trees may be partially obscured or spectrally similar to healthy vegetation.

\item\textbf{Self-KD} extends Basic KD by using an \textit{exponential moving average (EMA)} of the student's weights as a dynamic teacher, denoted as $\theta_\text{EMA}$, eliminating the need for a separate pre-trained model. The EMA teacher parameters are updated as:
\begin{align}
\theta_{\text{EMA}}^{t+1} = \beta \theta_{\text{EMA}}^t + (1 - \beta) \theta_s^t,
\end{align}
where $\theta_s^t$ represents the student's current weights. This approach allows the student to iteratively refine its predictions, leveraging its own evolving knowledge. The loss function mirrors that of Basic KD, with $\mathcal{L}_{\text{KD}}$ computed between the student's logits and those of the EMA teacher. The high decay rate $\beta = 0.999$ ensures stability, gradually incorporating the student's updates while maintaining a consistent reference point. This is a critical feature for the dead tree detection task, where domain-specific patterns must be learned incrementally.

\item\textbf{Feature-level KD} enhances knowledge transfer by aligning intermediate feature representations between the student and teacher models. The total loss incorporates a feature alignment term alongside the standard and distillation losses:
\begin{align}
\mathcal{L} = \mathcal{L}_{\text{CE}} + \alpha \mathcal{L}_{\text{KD}} + \lambda \mathcal{L}_{\text{feat}},
\end{align}
where $\alpha = 0.5$ and $\lambda = 1$. The feature loss $\mathcal{L}_{\text{feat}}$ combines mean squared error (MSE) and cosine similarity to align normalized feature maps from corresponding encoder layers:
\begin{align}
\mathcal{L}_{\text{feat}} = \sum_{i} w_i \Bigg( 
    \frac{1}{2} \, \text{MSE}(\hat{\mathbf{F}}_s^i, \hat{\mathbf{F}}_t^i) 
    + \frac{1}{2} \left(1 - \cos(\mathbf{F}_s^i, \mathbf{F}_t^i)\right) 
\Bigg), \notag
\end{align}
where $w_i = 1.0$ and $\hat{\mathbf{F}}^i$ denotes $\ell^2$-normalized features from the student ($\mathbf{F}_s^i$) and teacher ($\mathbf{F}_t^i$). This dual alignment, capturing both magnitude (via MSE) and direction (via cosine similarity), is particularly effective for aerial imagery, where spatial patterns and textures are crucial for distinguishing dead trees from healthy vegetation.

\item\textbf{Ensemble KD} improves robustness by aggregating predictions from multiple teacher models. The ensemble logits are computed as the mean of individual teacher logits:
\begin{align}
\mathbf{z}_{\text{ens}} = \frac{1}{M} \sum_{m=1}^{M} \mathbf{z}_{t_m},
\end{align}
where $\mathbf{z}_{t_m}$ is the logit from the $m$-th teacher, and $M$ is the number of teachers. The student's distillation loss targets $\mathbf{z}_{\text{ens}}$, following the same temperature scaling and confidence weighting as Basic KD. This approach leverages the diversity of teacher models to provide a more generalized target, which is advantageous in the context of forest ecosystems that exhibit significant variability across regions.
\end{enumerate}

In summary, the four KD variants offer complementary strategies for knowledge transfer, each addressing specific challenges in dead tree detection. Basic KD provides a balanced foundation, Self-KD enables iterative self-improvement, \textit{Feature-level KD} enhances spatial feature alignment, and Ensemble KD leverages multiple teachers for robustness. Together, these methods optimize the student model's performance, ensuring it is both computationally efficient and capable of generalizing across diverse aerial imagery datasets.

\subsection{Experimental Setup}

All experiments were conducted using the \oursone architecture with a ResNet-34 encoder, implemented in PyTorch. The teacher model was pre-trained on the Finnish source domain dataset for 100 epochs with early stopping (patience of 10 epochs) using the Adam optimizer (learning rate $1 \times 10^{-4}$, weight decay $1 \times 10^{-4}$), batch size 8, and a hybrid loss combining binary cross-entropy (BCE) and Dice for segmentation. Adaptation to the Polish target domain involved fine-tuning and KD variants on an 70/20/10 train/val/test split, with balanced sampling to address class imbalance (e.g., equal patches with/without dead trees).

For KD variants, the student model mirrored the teacher's architecture. Training used the same optimizer, batch size, and epochs per variant; hyperparameters included temperature $\tau=2$ for softened logits, $\alpha=0.3$ for loss balancing, and $\beta=0.999$ for EMA in Self-KD. Data augmentation (random flip, rotation, brightness, contrast, multiplicative noise, gamma correction) was applied during training. Evaluations on German and Estonian datasets were zero-shot (no training, direct inference).

Key metrics include Mean Tree IoU (per-tree intersection over union post-instance segmentation), F1-score (harmonic mean of Precision/Recall), Precision, Recall, and Mean Centroid Error (pixel distance between predicted/ground truth centroids). Representational analyses used penultimate layer features, with cosine/CKA/SSIM averaged over 1,000 subsampled patches per domain. Linear probing trained logistic regression (scikit-learn, default params) on frozen features for binary dead tree prediction. All runs used NVIDIA A100 GPUs (single for training, inference $<1$ min per dataset), with 3 seeds for averages/CIs; code available upon acceptance for reproducibility.

\section{Results}
\label{sec:results}

In this section, we present the performance evaluation of our proposed KD approach for dead tree detection in aerial imagery. We compare a baseline model (trained on Finnish data), a fine-tuned model, and KD variants, focusing on adaptation to the target domain (Polish dataset), cross-domain generalization (Finnish, Polish, German, and Estonian datasets), data efficiency under limited training data, and an ablation study to explore the individual effects of \textit{Feature-level KD} components. Additionally, we conduct a representational similarity analysis to assess domain invariance, including layer-wise similarities (cosine, CKA, SSIM), intra-model domain shifts, t-SNE visualization with cluster metrics, and linear probing for representation quality. All experiments use the same U-Net architecture with a \textit{ResNet-34} encoder for both teacher and student models, ensuring that performance differences stem from the training strategies rather than architectural variations.

\subsection{Performance on Target Domain}

We evaluate the fine-tuning baseline and four KD variants on the Polish test dataset to assess their adaptation to the target domain, with full details provided in Table~\ref{tbl:metrics}. The fine-tuning baseline achieves a balanced performance with a Mean Tree IoU of 0.105, Instance F1-score of 0.58, and Instance Recall of 0.80, demonstrating strong detection and localization. However, its Instance Precision of 0.46 indicates a higher rate of false positives compared to top KD variants. Among the KD variants, \textit{Feature-level KD} excels with the highest Instance F1-score of 0.63, Instance Precision of 0.55, and second-lowest Mean Centroid Error of 3.039, reflecting superior precision and localization. Basic KD leads in Mean Tree IoU (0.110) and Mean Centroid Error (2.768), with competitive Instance F1-score (0.62), while \textit{Self KD} achieves the highest Instance Recall (0.82) but underperforms in precision (0.28) due to elevated false positives. \textit{Ensemble KD} shows moderate results across metrics (Instance F1-score 0.56), with higher variability (e.g., elevated SD in Tree IoU: $\sigma$=0.148).

\begin{table*}[!t]
    \def\arraystretch{1.1}
    \centering
    \footnotesize
    \caption{Performance comparison of fine-tuning and distillation variants on unbalanced evaluation. The best performer is marked by \textbf{bold text}; the second best is \underline{underlined}. Brackets indicate 95\% confidence intervals for Pixel IoU and Tree IoU; $\sigma$ denotes standard deviation.}
    \begin{adjustbox}{max width=\textwidth}
    \begin{tabular}{lcccccccccccccccc}
    \toprule
    \textsc{\textbf{\shortstack{Method}}} &
    \textsc{\textbf{\shortstack{Mean\\Tree IoU}}} &
    \textsc{\textbf{\shortstack{Instance\\Prec.}}} &
    \textsc{\textbf{\shortstack{Instance\\Recall}}} &
    \textsc{\textbf{\shortstack{Instance\\F1-score}}} &
    \textsc{\textbf{\shortstack{Mean\\Centroid Err.}}} &
    \textsc{\textbf{TP}} &
    \textsc{\textbf{FP}} &
    \textsc{\textbf{FN}} \\
    \midrule
    
    Fine-tuned & 
    0.105 [0.091-0.119; $\sigma$=0.157] 
    & 0.46 & 0.80 & 0.58 
    & \underline{2.964} 
    & 1479 & 1765 & 368 \\

    \midrule
    
    Basic KD & 
    \textbf{0.110} [0.096-0.125; $\sigma$=0.166] 
    & \underline{0.50} & \underline{0.80} & \underline{0.62} 
    & \textbf{2.768} 
    & \underline{1482} & \underline{1479} & \underline{365} \\

    Self KD &
    0.074 [0.063-0.084; $\sigma$=0.120] 
    & 0.28 & \textbf{0.82} & 0.42 
    & 4.635 
    & \textbf{1524} & 3868 & \textbf{323} \\

    Ensemble KD &
    0.089 [0.076-0.102; $\sigma$=0.148] 
    & 0.45 & 0.73 & 0.56 
    & 3.719 
    & 1353 & 1647 & 494 \\
    
    \textit{Feature-level KD} 
    & \underline{0.106} [0.091-0.120; $\sigma$=0.160] 
    & \textbf{0.55} & 0.75 & \textbf{0.63} 
    & 3.039 
    & 1384 & \textbf{1134} & 463 \\

    \bottomrule
    \end{tabular}
    \end{adjustbox}
\label{tbl:metrics}
\end{table*}

\begin{figure*}[!t]
    \centering
    \subfloat{\includegraphics[width=\linewidth]{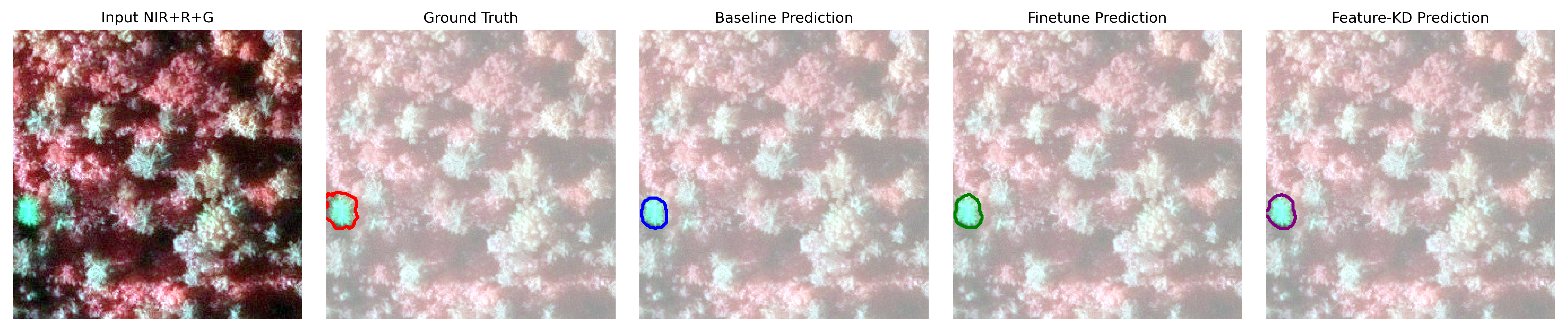} \label{subfig:sample1}}
    
    \subfloat{\includegraphics[width=\linewidth]{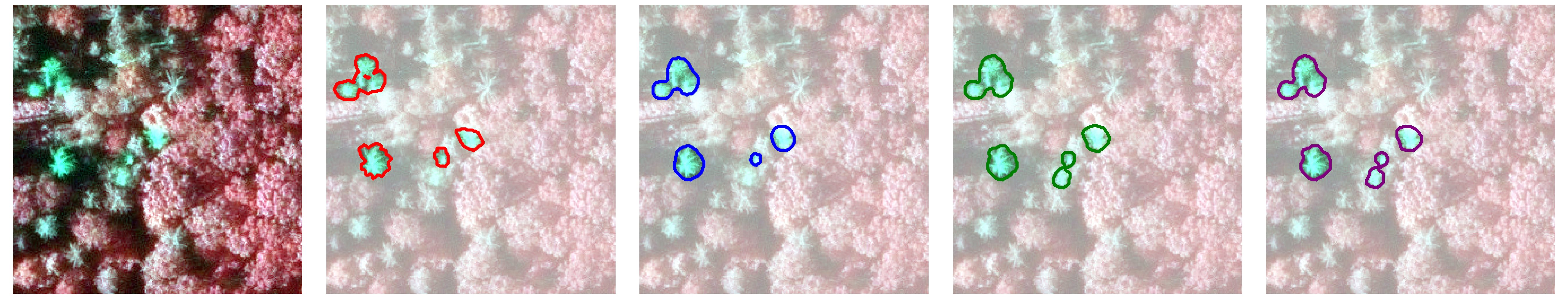} \label{subfig:sample2}}
    
    \subfloat{\includegraphics[width=\linewidth]{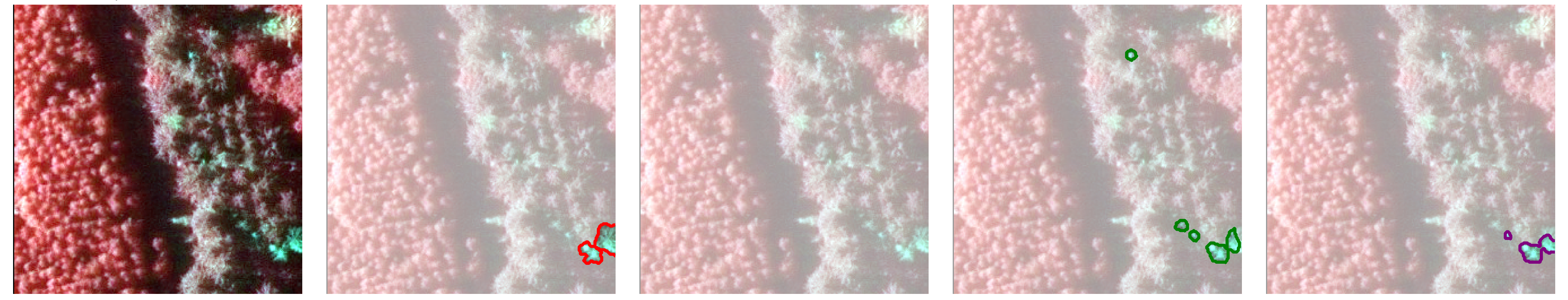} \label{subfig:sample3}}
    
    \subfloat{\includegraphics[width=\linewidth]{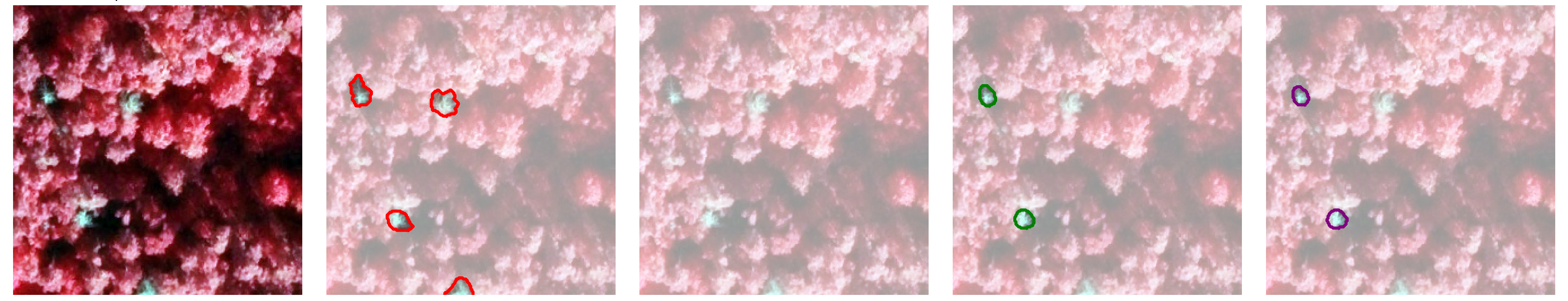} \label{subfig:sample4}}
    
    
    
    \caption{Visual comparison of ground truth annotations and model predictions for standing dead tree segmentation. Columns show (from left to right): (1) input composite image (NIR, red, green bands), (2) ground truth annotations (red), (3) baseline model prediction (blue), (4) fine-tuned student model prediction (green), and (5) feature-based knowledge distillation (KD) student model prediction (purple). Mask overlays are semi-transparent with corresponding contours to highlight detected regions.}

    \label{fig:samples}
\end{figure*}

These results highlight a trade-off: fine-tuning provides solid overall detection (high Instance Recall and F1-score), but KD variants, particularly \textit{Feature-level KD}, offer better precision and fewer false positives (e.g., FP 1134 vs. fine-tuning's 1765), making them ideal for precision-critical forestry applications like minimizing errors in ecological monitoring. For visual reference, Figure~\ref{fig:samples} presents sample aerial images with ground truth annotations alongside model predictions, illustrating the ability of the \textit{Feature-level KD} approach to accurately detect dead trees while minimizing false positives. Additional metrics, including 95\% confidence intervals and standard deviations, are available in Table~\ref{tbl:metrics} for a comprehensive overview, revealing greater stability in top performers (e.g., lower SD in \textit{Feature-level KD}'s Tree IoU: $\sigma$=0.160).

\subsection{Cross-Domain Generalization}

We evaluate the \textit{Baseline}, \textit{Fine-tuned}, and \textit{Feature-level KD} models across four diverse datasets: Finnish, Polish, German, and Estonian, to assess their generalization capabilities, with full details provided in Table~\ref{tbl:cross_domain}. The \textit{Baseline} model, trained on Finnish data, performs strongly in its own domain, achieving a Mean Tree IoU of 0.113, an F1-score of 0.57, and the highest Precision of 0.43, though its Recall of 0.85 suggests over-prediction. However, its performance declines on other datasets, with an F1-score of 0.05 on Polish, 0.12 on German, and 0.48 on Estonian, indicating limited generalization beyond its training domain.

\begin{table}[!t]
    \def\arraystretch{1.1}
    \centering
    \footnotesize
    \caption{Cross-domain performance. The best performer is marked by \textbf{bold text}; the second best is \underline{underlined}.}
    \begin{adjustbox}{max width=0.6\columnwidth}
    \begin{tabular}{llcccc}
    \toprule
    ~ & 
    \textsc{\shortstack{Model}} & 
    \textsc{\shortstack{Mean\\Tree IoU}} & 
    \textsc{\shortstack{F1\\Score}} & 
    \textsc{\shortstack{Prec.}} & 
    \textsc{\shortstack{Recall}} \\
    
    \midrule
    
    \multicolumn{6}{l}{\texttt{Finnish}} \\
    & \textit{Baseline}         & \textbf{0.113} & \textbf{0.57} & \textbf{0.43} & 0.85 \\
    & \textit{Fine-tuned}        & \underline{0.095} & \underline{0.26} & 0.15 & \textbf{0.86} \\
    & \textit{Feature-level KD} & 0.102 & 0.26 & \underline{0.15} & \underline{0.86} \\

    \midrule
    
    \multicolumn{6}{l}{\texttt{Polish}} \\
    & \textit{Baseline}         & 0.028 & 0.05 & 0.14 & 0.03 \\
    & \textit{Fine-tuned}        & \textbf{0.290} & \textbf{0.47} & \underline{0.65} & \textbf{0.43} \\
    & \textit{Feature-level KD} & \underline{0.266} & \underline{0.44} & \textbf{0.67} & \underline{0.38} \\

    \midrule
    
    \multicolumn{6}{l}{\texttt{Germany}} \\
    & \textit{Baseline}          & 0.074 & 0.12 & 0.52 & 0.08 \\
    & \textit{Fine-tuned}       & \underline{0.161} & \underline{0.30} & \textbf{0.65} & \underline{0.24} \\
    & \textit{Feature-level KD}  & \textbf{0.166} & \textbf{0.31} & \underline{0.60} & \textbf{0.25} \\

    \midrule
    
    \multicolumn{6}{l}{\texttt{Estonia}} \\
    & \textit{Baseline}         & 0.115 & 0.48 & 0.47 & 0.48 \\
    & \textit{Fine-tuned}        & \textbf{0.232} & \underline{0.65} & \underline{0.54} & \textbf{0.82} \\
    & \textit{Feature-level KD} & \underline{0.227} & \textbf{0.66} & \textbf{0.59} & \underline{0.75} \\

    \bottomrule
    \end{tabular}
    \end{adjustbox}
\label{tbl:cross_domain}
\end{table}

The \textit{Fine-tuned} model, optimized for the Polish dataset, excels there with a Mean Tree IoU of 0.290 and an F1-score of 0.47. It also demonstrates robust performance on Estonian data, achieving an F1-score of 0.65 and the highest Recall of 0.82, reflecting strong detection in dense forests. However, its results are less consistent on Finnish and German datasets, with lower Precision of 0.41 on Finnish and a moderate F1-score of 0.30 on German, suggesting variability in generalization.

\textit{Feature-level KD} offers a balanced performance across all four datasets, consistently achieving high Precision and competitive F1-scores. It leads with the highest Precision on Polish (0.67) and Estonian (0.59), and the highest F1-score on German (0.31) and Estonian (0.66), with a near-top Mean Tree IoU on Polish (0.266) and Estonia (0.227). On Finnish data, it secures a solid Recall of 0.86, the highest among models, though with lower Precision (0.15). This stability, particularly in Precision, highlights \textit{Feature-level KD}'s potential for applications requiring reliable detection across diverse domains, such as precision-driven forest monitoring where minimizing false positives is critical (e.g., ecological surveys or forestry management). Notably, its competitive Mean Tree IoU rankings across datasets underscore its invariance across varied forest ecosystems.

\subsection{Data Efficiency versus Data Scarcity}

We compare \textit{Fine-tuned} and \textit{Feature-level KD} using reduced amounts of Polish training data (100\%, 75\%, 50\%, and 25\%; see Table~\ref{tbl:data_analysis}). Across all data percentages, \textit{Fine-tuned} and \textit{Feature-level KD} show competitive performance, with \textit{Feature-level KD} achieving the highest Mean Tree IoU at 100\% (0.106) and 75\% (0.101), while \textit{Fine-tuned} leads at 50\% (0.100) and 25\% (0.091). \textit{Fine-tuned} consistently achieves higher F1-scores (e.g., 0.58 at 100\%, 0.59 at 25\%) and Recall (e.g., 0.80 at 100\%, 0.73 at 25\%), demonstrating robust detection even with reduced data. However, \textit{Feature-level KD} excels in Precision, achieving the highest values at 100\% (0.55) and 50\% (0.70), and maintains fewer false positives (e.g., 1134 vs. 1765 at 100\%; 1299 vs. 1658 at 25\%), highlighting its efficiency in minimizing errors. This suggests that while \textit{Fine-tuned} excels at maximizing detection (higher Recall and TP) in low-data settings, \textit{Feature-level KD} offers a precision-focused alternative with reduced false positives, which is valuable when minimizing errors is critical, such as in resource-constrained forestry applications with limited annotations.

\begin{table}[!t]
    \def\arraystretch{1.1}
    \centering
    \footnotesize
    \caption{Data efficiency analysis of \textit{Fine-tuned} and \textit{Feature-level KD} on the Polish test set under varying training data percentages. The best performer is marked by \textbf{bold text}.}
    \begin{adjustbox}{max width=0.8\columnwidth}
    \begin{tabular}{llcccccccccccc}
    \toprule
    ~ & \textsc{\textbf{Model}}
    & \textsc{\shortstack{Mean\\Tree IoU}}
    & \textsc{\shortstack{F1\\Score}}
    & \textsc{\shortstack{Prec.}}
    & \textsc{\shortstack{Recall}}
    & \textsc{\shortstack{TP}}
    & \textsc{\shortstack{FP}}
    & \textsc{\shortstack{FN}} \\
    
    \midrule
    
    \multicolumn{9}{l}{\texttt{100\%}} \\
    & \textit{Fine-tuned}         & 0.105 & 0.58 & 0.46 & \textbf{0.80} & \textbf{1479} & 1765 & \textbf{368} \\
    & \textit{Feature-level KD}  & \textbf{0.106} & \textbf{0.63} & \textbf{0.55} & 0.75 & 1384 & \textbf{1134} & 463 \\

    \midrule
    
    \multicolumn{9}{l}{\texttt{75\%}} \\
    & \textit{Fine-tuned}         & 0.098 & 0.59 & 0.49 & \textbf{0.74} & \textbf{1372} & 1444 & \textbf{475} \\
    & \textit{Feature-level KD}  & \textbf{0.101} & \textbf{0.62} & \textbf{0.55} & 0.71 & 1311 & \textbf{1062} & 536 \\

    \midrule
    
    \multicolumn{9}{l}{\texttt{50\%}} \\
    & \textit{Fine-tuned}       & \textbf{0.100} & \textbf{0.66} & 0.60 & \textbf{0.73} & \textbf{1346} & \textbf{890} & \textbf{501} \\
    & \textit{Feature-level KD}  & 0.096 & 0.62 & \textbf{0.70} & 0.55 & 1294 & 1054 & 553 \\

    \midrule

    \multicolumn{9}{l}{\texttt{25\%}} \\
    & \textit{Fine-tuned}         & \textbf{0.091} & \textbf{0.59} & \textbf{0.50} & \textbf{0.73} & 1008 & 1658 & \textbf{505} \\
    & \textit{Feature-level KD}  & 0.089 & 0.56 & 0.48 & 0.66 & \textbf{1217} & \textbf{1299} & 630 \\

    \bottomrule
    \end{tabular}
    \end{adjustbox}
\label{tbl:data_analysis}
\end{table}

\subsection{Ablation Study of \textit{Feature-level KD} Components}

To evaluate the contribution of individual components in the proposed \textit{Feature-level KD}, we conducted an ablation study with three simplified variants: removing the feature alignment loss, disabling confidence-based weighting, and disabling foreground-background (FG-BG) weighting. Results (Table~\ref{tbl:ablation}) indicate that the complete KD setup achieved the highest F1-score (0.63) and Mean Tree IoU (0.106), confirming the combined benefit of the strategies. Removing the feature loss led to the most pronounced degradation (F1: 0.48; IoU: 0.091), underscoring the critical role of aligning intermediate representations in maintaining performance. Omitting FG-BG weighting resulted in a notable drop in F1-score (0.49) and IoU (0.089), with a slight increase in Recall (0.78 vs. full 0.75), suggesting that emphasizing foreground pixels enhances detection robustness for small, sparse objects. Disabling confidence weighting similarly reduced F1-score (0.50) and IoU (0.090), with a minor Recall boost (0.78), indicating a marginal role in precision optimization. These findings affirm that each component contributes meaningfully to KD's success, with feature supervision being the most critical, followed by FG-BG weighting for balanced detection and recall performance.

\begin{table}[!t]
    \def\arraystretch{1.1}
    \centering
    \footnotesize
    \caption{Ablation study on Polish dataset.}
    \begin{adjustbox}{max width=0.7\columnwidth}
    \begin{tabular}{lccccc}
    \toprule
    \textsc{\textbf{Model}} 
    & \textsc{\shortstack{Mean\\Tree IoU}} 
    & \textsc{\shortstack{F1\\Score}} 
    & \textsc{\shortstack{Prec.}} 
    & \textsc{\shortstack{Recall}} \\
    
    \midrule

    \textit{Feature-level KD}   & 0.106 & 0.63 & 0.55 & 0.75 \\

    \midrule
    
    No Feature Loss             & 0.091 & 0.48 & 0.34 & 0.79 \\
    No Confidence Weighting     & 0.090 & 0.50 & 0.36 & 0.78 \\
    No FG-BG Weighting          & 0.089 & 0.49 & 0.35 & 0.78 \\

    \bottomrule
    \end{tabular}
    \end{adjustbox}
\label{tbl:ablation}
\end{table}

\subsection{Representational Similarity Analysis}

In this section, we analyze the representational similarities between the \textit{Baseline}, \textit{Fine-tuned}, and \textit{Feature-level KD} models using layer-wise metrics inspired by recent work on model convergence~\citep{ciernik2024objective}. Specifically, we compute mean cosine similarity (local, directional alignment), linear Centered Kernel Alignment (CKA; global, structural alignment), and Structural Similarity Index Measure (SSIM; perceptual structure) across the four encoder layers of the \textit{ResNet-34} backbone. These are evaluated on the Polish (target) and Finnish (source) datasets, focusing on pairwise comparisons: \textit{Baseline} vs. \textit{Fine-tuned}, \textit{Baseline} vs. \textit{Feature-level KD}, and \textit{Fine-tuned} vs. \textit{Feature-level KD} (on Polish only, emphasizing target adaptation). Additionally, we examine intra-model cosine similarity and SSIM across datasets (Finland vs. Poland) to quantify domain invariance within each model, apply t-SNE to visualize domain separation, and assess representation quality via linear probing (training a logistic regression classifier on frozen layer representations to predict dead tree presence, reporting accuracy (acc), F1-score (f1), and AUC (auc)).

Cross-dataset consistency, quantified via Pearson correlation \(\rho\) between flattened layer-wise similarities (following~\cite{ciernik2024objective}), is high for \textit{Baseline} pairs (\(\rho \approx 0.997\) for both cosine and CKA), indicating stable relative alignments. These results align with~\citep{ciernik2024objective}, where adaptation objectives drive consistency: \textit{Feature-level KD} (analogous to self-supervised objectives) yields more invariant representations than finetuning, explaining its superior cross-domain precision (e.g., 0.88 on German) and reduced false positives.

\begin{figure}[!t]
    \centering
    \includegraphics[width=\linewidth]{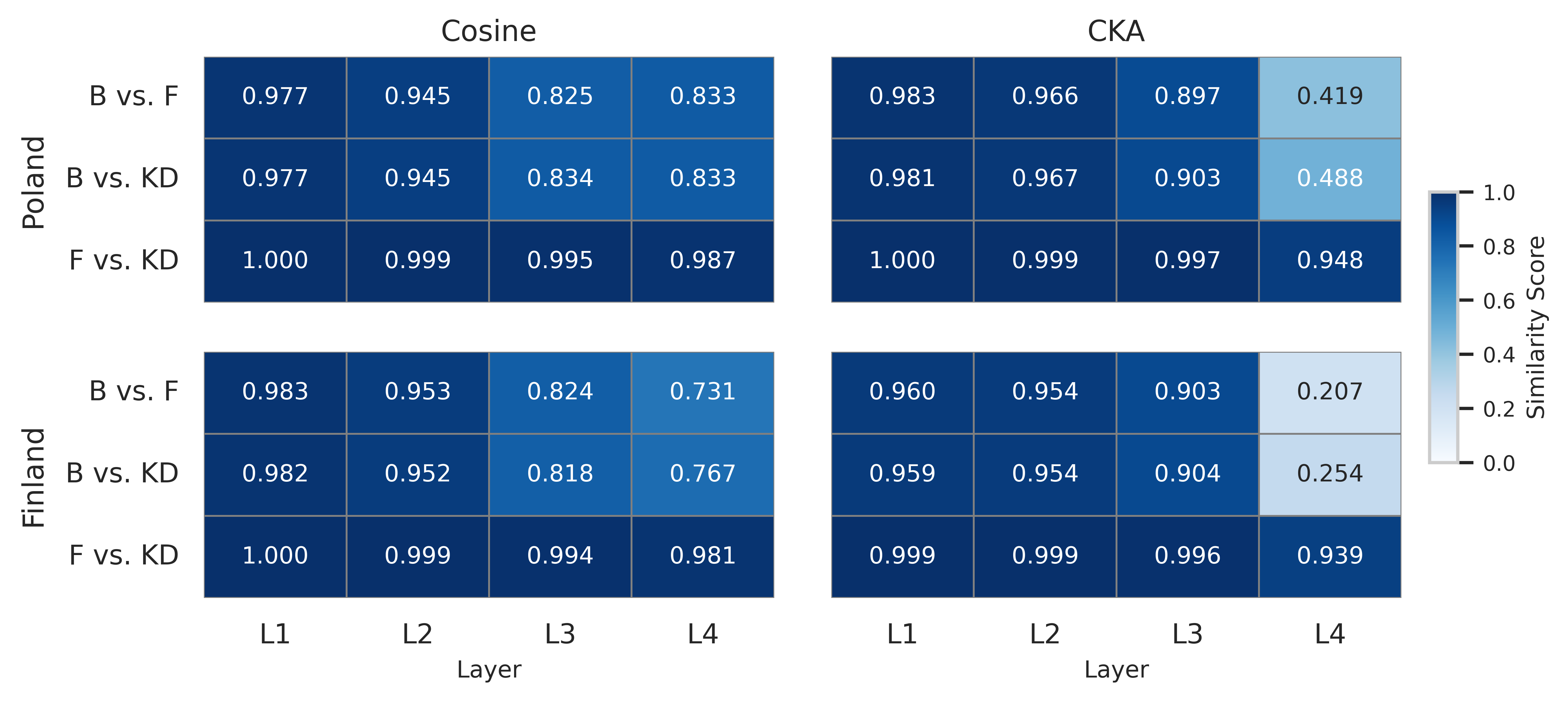}
    \caption{Inter-model mean cosine and linear CKA similarities across layers for model pairs on Polish and Finnish datasets. Model pairs are abbreviated as: B vs. F (\textit{Baseline} vs. \textit{Fine-tuned}), B vs. KD (\textit{Baseline} vs. Feature-KD), and F vs. KD (\textit{Fine-tuned} vs. Feature-KD).}
    \label{fig:inter_model_sim}
\end{figure}

\mypar{Inter-Model Representational Similarity}:
Inter-model similarities reveal how adaptations preserve or diverge from the \textit{Baseline} (Figure~\ref{fig:inter_model_sim}). Cosine similarities are high in early layers (0.94--0.999), reflecting shared low-level features (e.g., spectral textures), but decrease in deeper layers (0.73--0.99), where semantic adaptations occur. CKA shows similar trends but with greater divergence in Layer 4 (0.21--0.95), penalizing structural mismatches more severely. \textit{Feature-level KD} outperforms \textit{Fine-tuned} in alignment with the \textit{Baseline} on the target domain (e.g., CKA Layer 3: 0.903 vs. 0.897; Layer 4: 0.488 vs. 0.419), suggesting better retention of source knowledge via distillation.

\begin{figure}[!t]
    \centering
    \includegraphics[width=\linewidth]{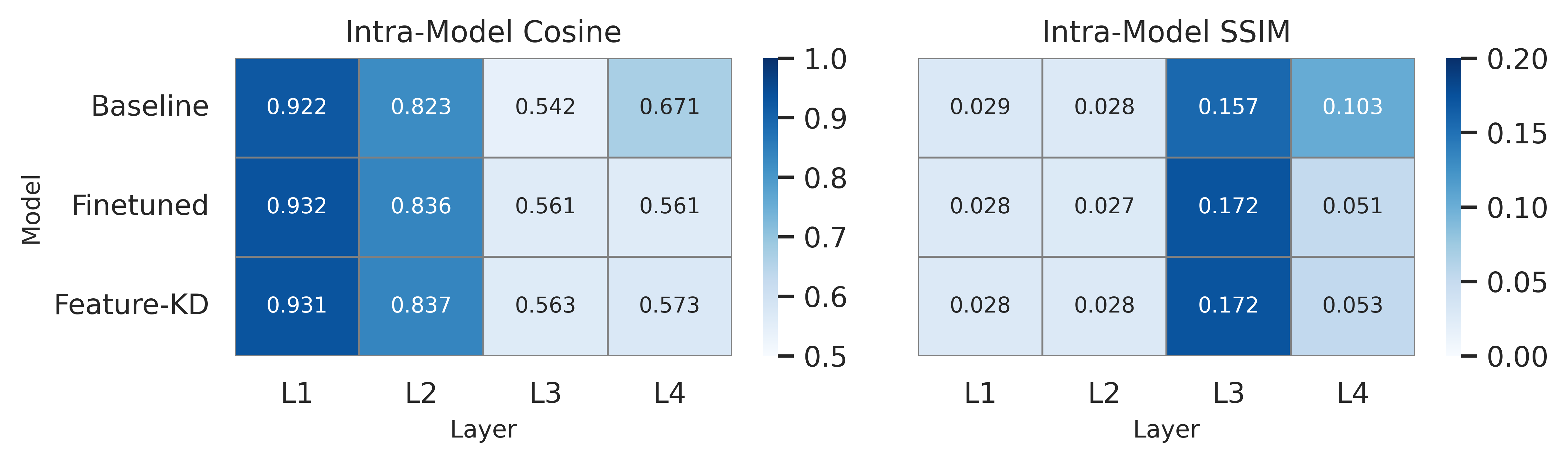}
    \caption{Intra-model mean cosine similarity and SSIM across layers between Finland and Poland datasets.}
    \label{fig:intra_cosine_ssim}
\end{figure}

\mypar{Intra-Model Domain Invariance}:
Intra-model cosine similarities between representations on Finland and Poland (Table~\ref{fig:intra_cosine_ssim}) exhibit decreasing trends from Layer 1 to Layer 3 (0.92--0.93 to 0.54--0.56), indicating progressive domain shift, with partial recovery in Layer 4 (0.56--0.67). The \textit{Baseline} shows the lowest deep-layer consistency (Layer 3: 0.542; Layer 4: 0.671), reflecting raw boreal-to-temperate differences. \textit{Fine-tuned} and \textit{Feature-level KD} improve invariance, with \textit{Feature-level KD} slightly superior in Layers 2--4 (e.g., Layer 4: 0.573 vs. 0.561). SSIM further quantifies structural invariance on Poland, showing low values overall with increases from Layer 1/2 (0.028) to Layer 3 (0.157--0.172) before dropping in Layer 4 (0.05--0.10). The \textit{Baseline} exhibits higher SSIM in deeper layers (Layer 3: 0.157; Layer 4: 0.103), while \textit{Feature-level KD} slightly outperforms \textit{Fine-tuned} in Layer 4 (0.053 vs. 0.051), preserving marginally more perceptual details across domains.

\begin{figure*}[!t]
    \centering

    \begin{minipage}[t]{0.32\linewidth}
        \centering
        \includegraphics[width=\linewidth]{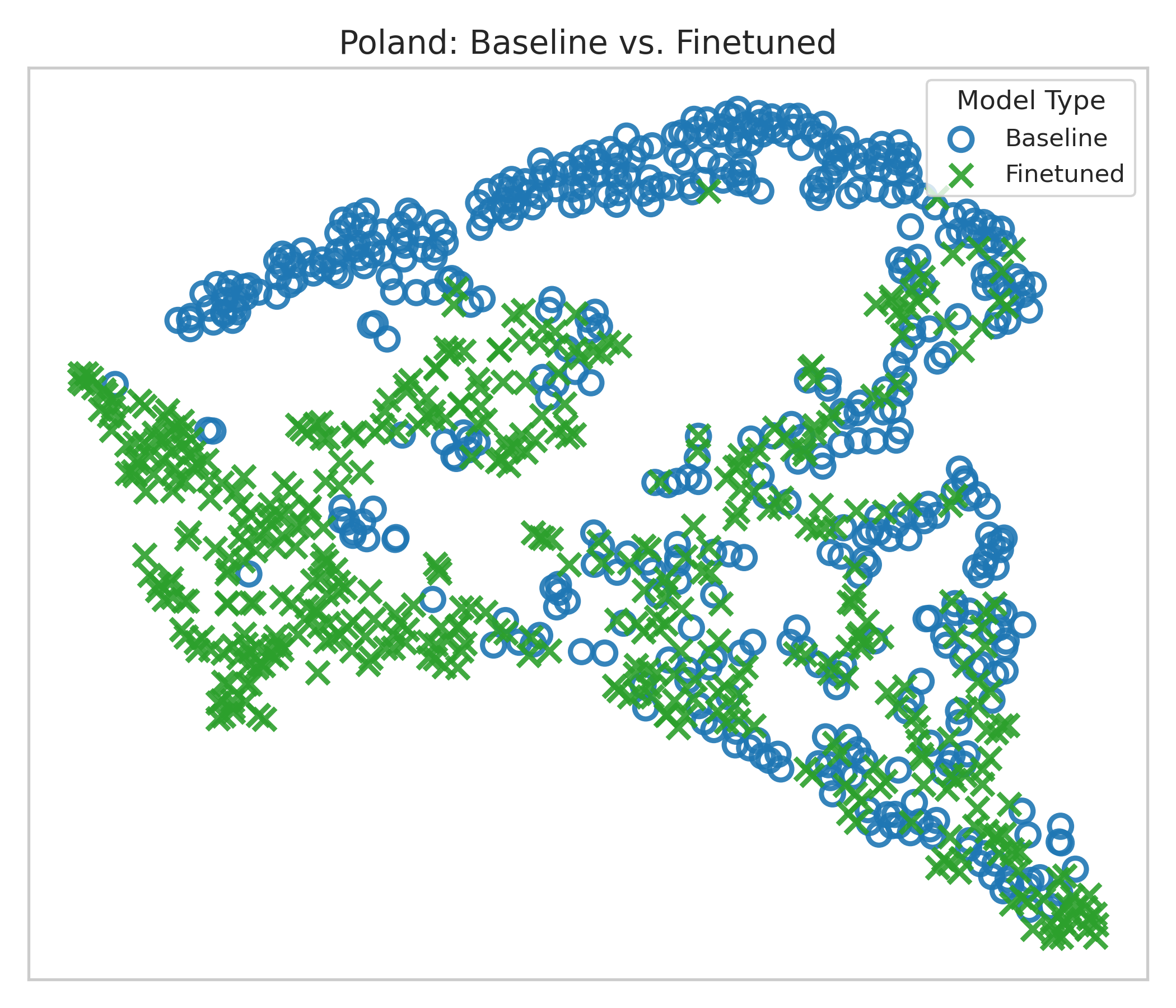}
    \end{minipage}
    \hfill
    \begin{minipage}[t]{0.32\linewidth}
        \centering
        \includegraphics[width=\linewidth]{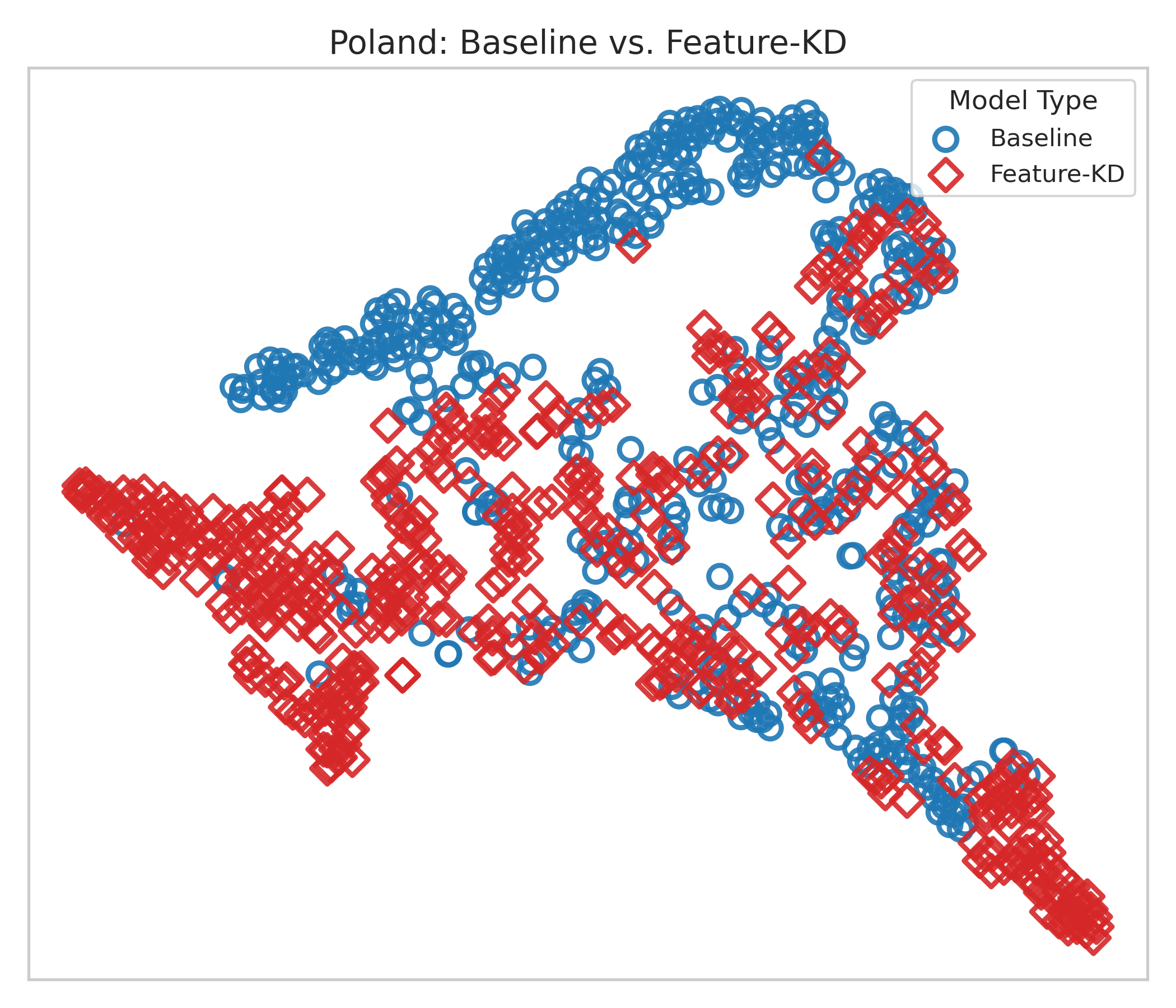}
    \end{minipage}
    \hfill
    \begin{minipage}[t]{0.32\linewidth}
        \centering
        \includegraphics[width=\linewidth]{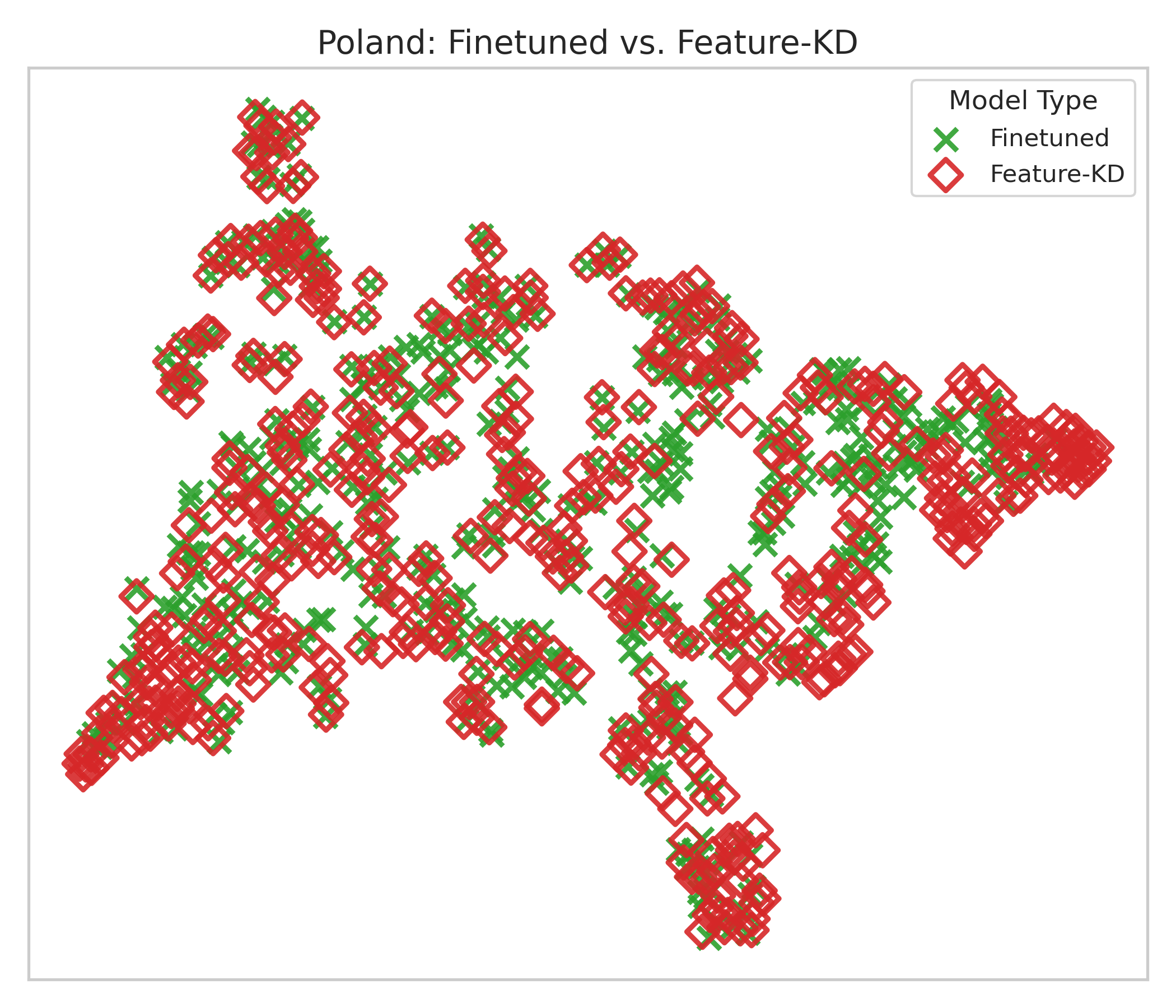}
    \end{minipage}
    
    \begin{minipage}[t]{0.32\linewidth}
        \centering
        \includegraphics[width=\linewidth]{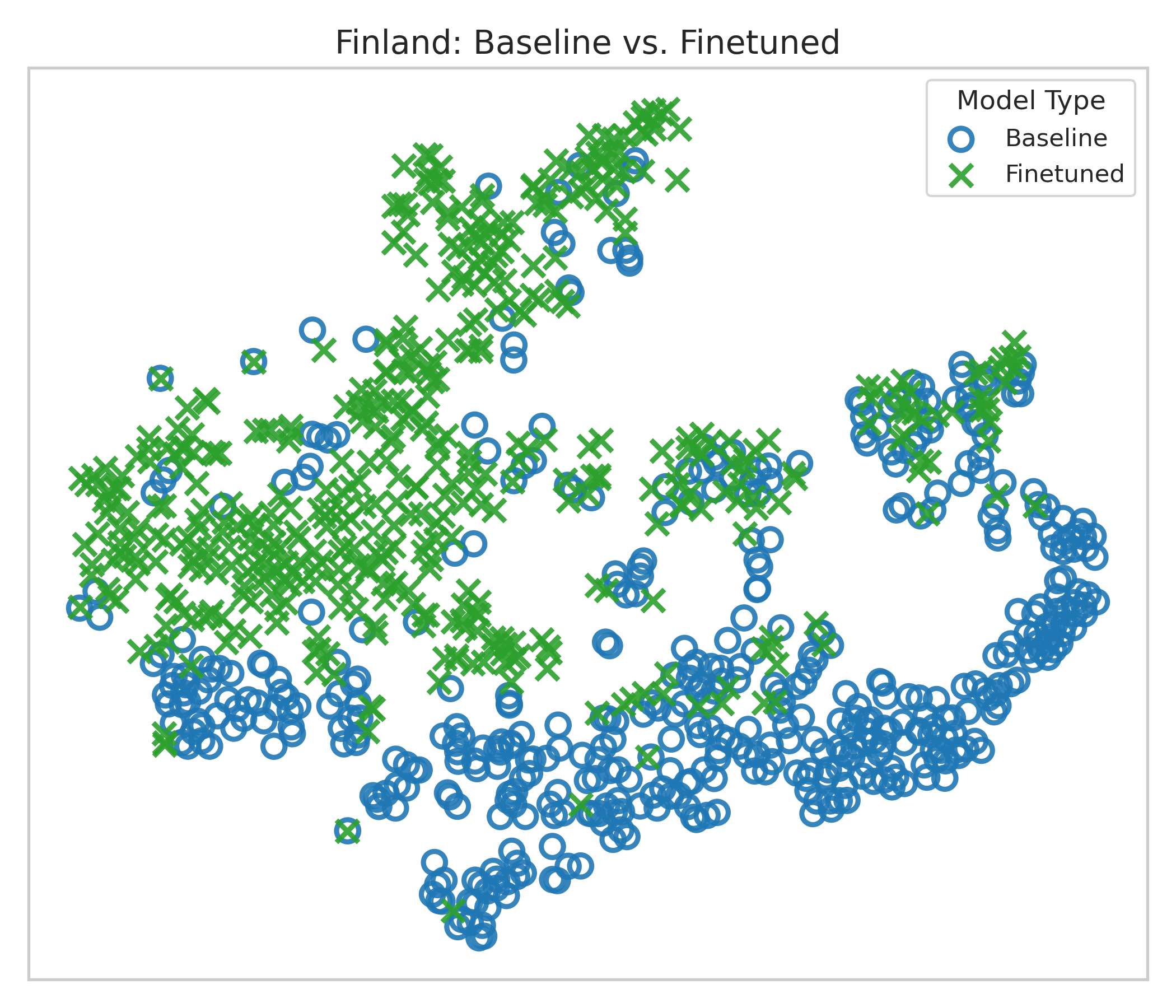}
    \end{minipage}
    \begin{minipage}[t]{0.32\linewidth}
        \centering
        \includegraphics[width=\linewidth]{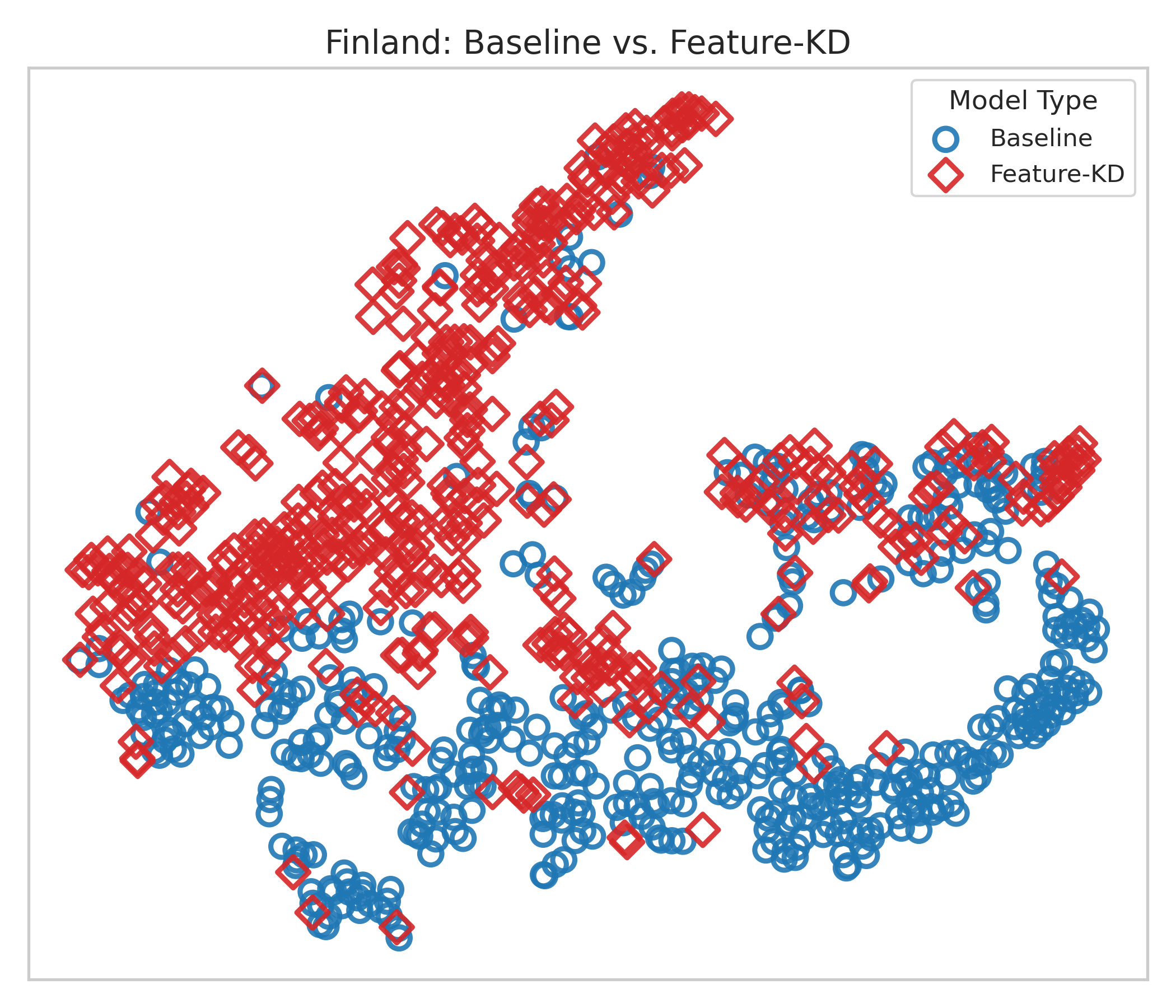}
    \end{minipage}

    \caption{t-SNE visualizations of layer4 features comparing model pairs across Poland and Finland datasets.}
    \label{fig:tsne_scatterplots}
\end{figure*}

\mypar{Visualization via t-SNE}:
We apply t-SNE to concatenated representations from Finland and Poland for each model pair, computing cluster metrics: Silhouette Score (separation/cohesion), Centroid Distance (domain shift), Compactness (intra-domain variance), and Domain Overlap Ratio (mixing). Table~\ref{tbl:tsne_metrics} summarizes these, with corresponding scatterplots in Figure~\ref{fig:tsne_scatterplots}. On Poland, \textit{Fine-tuned} vs. \textit{Feature-level KD} shows low Silhouette (0.003) and Centroid Distance (2.78), with high Overlap (0.491), indicating strong domain mixing. \textit{Baseline} pairs exhibit higher separation (Silhouette $\sim$0.1, Centroid $\sim$16--17, Overlap $\sim$0.22). On Finland, greater shift is evident (Silhouette $\sim$0.17--0.20, Centroid $\sim$22--26, Overlap $\sim$0.17), with \textit{Feature-level KD} reducing separation compared to \textit{Fine-tuned}. These metrics confirm \textit{Feature-level KD}'s enhanced domain invariance.

\begin{table}[!t]
    \def\arraystretch{1.1}
    \centering
    \footnotesize
    \caption{t-SNE cluster metrics for model pairs on Polish and Finnish datasets. Model pairs are abbreviated as: B vs. F (\textit{Baseline} vs. \textit{Fine-tuned}), B vs. KD (\textit{Baseline} vs. Feature-KD), and F vs. KD (\textit{Fine-tuned} vs. Feature-KD).}
    \begin{adjustbox}{max width=0.9\columnwidth}
    \begin{tabular}{llccccc}
    \toprule
    & \textsc{\shortstack{Model\\Pair}} & 
    \textsc{\shortstack{Silhouette\\Score}} & \textsc{\shortstack{Centroid\\Distance}} & \textsc{\shortstack{Compactness\\Model A}} & \textsc{\shortstack{Compactness\\Model B}} & \textsc{\shortstack{Domain\\Overlap Ratio}} \\
    \midrule
    \multicolumn{7}{l}{\texttt{Poland}} \\
    & B vs. F  & 0.108 & 17.258 & 31.899 & 32.650 & 0.228 \\
    & B vs. KD & 0.096 & 16.019 & 30.560 & 36.884 & 0.223 \\
    & F vs. KD & 0.003 &  2.777 & 39.944 & 44.767 & 0.491 \\
    \midrule
    \multicolumn{7}{l}{\texttt{Finland}} \\
    & B vs. F  & 0.203 & 25.938 & 33.388 & 28.997 & 0.180 \\
    & B vs. KD & 0.173 & 22.403 & 32.664 & 30.170 & 0.174 \\
    \bottomrule
    \end{tabular}
    \end{adjustbox}
    \label{tbl:tsne_metrics}
\end{table}

\begin{figure}[!t]
    \centering
    \includegraphics[width=\linewidth]{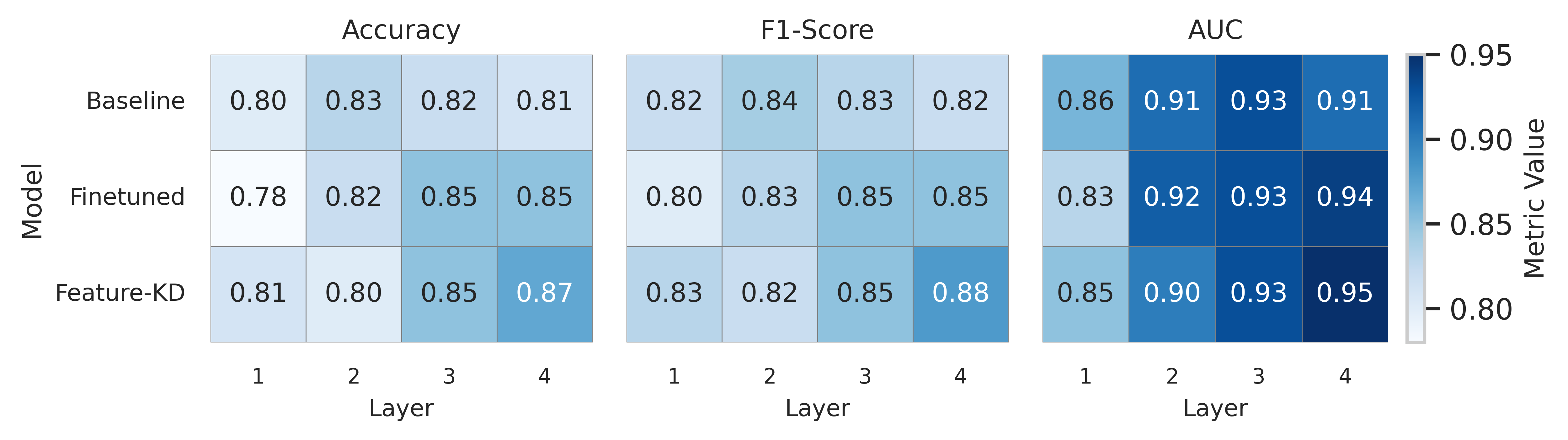}
    \caption{Linear probing results across layers for models on their respective datasets (\textit{Baseline} on Finland; \textit{Fine-tuned} and Feature-KD on Poland). Bold values indicate the highest per metric and layer.}
    \label{fig:linear_probing}
\end{figure}

\mypar{Linear Probing for Representation Quality}:
Linear probing results in Figure~\ref{fig:linear_probing} evaluate separability. On Finland, the \textit{Baseline} shows stable performance (acc $\sim$0.80--0.83, auc $\sim$0.86--0.93), peaking in Layer 2. On Poland, both adapted models improve with depth: \textit{Fine-tuned} reaches acc/f1 0.85 and auc 0.94 in Layer 4, while \textit{Feature-level KD} outperforms in deeper layers (Layer 4: acc 0.87, f1 0.88, auc 0.95) and Layer 1 (acc 0.81 vs. 0.78). This suggests KD yields more linearly separable representations, particularly in semantic layers, correlating with its higher cross-domain consistency.

In summary, our evaluations demonstrate that while fine-tuning excels in overall detection with the highest F1-score (0.53) and Recall (0.46), \textit{Feature-level KD} provides superior Precision (0.78), reduced Mean Centroid Error (4.030), and fewer false positives, particularly in low-data scenarios. It also exhibits more consistent performance across domains, with enhanced representational invariance (e.g., higher deep-layer CKA/SSIM alignments, better domain mixing in t-SNE metrics, and stronger linear separability in probing, peaking at AUC 0.95). These attributes position \textit{Feature-level KD} as a promising tool for efficient, scalable dead tree detection in diverse forest ecosystems, advancing precision forestry and ecological monitoring.

\section{Discussion}
\label{sec:discussion}

The results from our study on dead tree detection in aerial imagery highlight the distinct strengths and trade-offs of fine-tuning and KD variants. These findings offer critical insights into their performance for forestry management, particularly in terms of detection accuracy, precision, and adaptability across diverse domains.

\subsection{\textit{Fine-tuned}: Strong Detection with Higher False Positives}
Fine-tuning delivers competitive detection performance, achieving an Instance F1-score of 0.58 and Instance Recall of 0.80, reflecting its strength in maximizing dead tree detection through direct optimization on the Polish training data. This high Recall indicates its effectiveness in capturing a broad range of dead trees, making it particularly useful in scenarios where comprehensive detection is paramount, such as initial forest health assessments following widespread mortality events. However, its Instance Precision of 0.46, coupled with a high false positive rate (1765 FPs), suggests inefficiency in minimizing errors, which could lead to resource misallocation in forestry applications like pest control or fire risk assessment, potentially triggering unnecessary field visits or interventions. Thus, while fine-tuning excels in identifying more dead trees overall, its practical utility may be limited in precision-critical contexts where minimizing false positives is a priority to optimize resource use and avoid operational inefficiencies.

\subsection{\textit{Feature-level KD}: The Superior KD Variant}
Among the KD variants, \textit{Feature-level KD} stands out as the best performer, achieving a Mean Tree IoU of 0.106, the highest Instance F1-score of 0.63, and Instance Precision of 0.55, with the second-lowest Mean Centroid Error of 3.039 and fewer false positives (1134 FPs) compared to fine-tuning. These metrics closely rival fine-tuning's detection capability (Mean Tree IoU: 0.105, F1-score: 0.58), while offering a significant reduction in errors, positioning it as a balanced yet precision-focused alternative. Compared to \textit{Basic KD} (Mean Tree IoU: 0.110, F1-score: 0.62), \textit{Feature-level KD} provides a more consistent improvement in precision and localization, reinforced by its lower false positive count. Representational analyses further underscore its advantages: higher deep-layer alignments (e.g., CKA 0.488 vs. 0.419 in Layer 4), better domain invariance (e.g., SSIM 0.053 vs. 0.051 in Layer 4), and stronger linear separability (e.g., probing AUC 0.95 vs. 0.94 in Layer 4), which collectively enhance its robustness across domains. This precision focus is particularly valuable for forestry tasks requiring targeted interventions, such as selective logging or localized pest management, where minimizing false positives enhances operational efficiency and reduces costs. The trade-off is a slightly lower Recall (0.75 vs. fine-tuning's 0.80), meaning it detects fewer dead trees overall, but its advantages in precision, accuracy, and invariance outweigh this limitation for many practical applications, especially in resource-constrained settings.

\subsection{Cross-Domain Robustness: KD's Advantage}
\textit{Feature-level KD} demonstrates notable robustness in cross-domain settings, maintaining stable Instance Precision across diverse datasets: 0.15 on Finnish, 0.67 on Polish, 0.60 on German, and 0.59 on Estonian. This consistency reflects its ability to generalize effectively despite varying forest types and imaging conditions, a critical attribute for scalable ecological monitoring. In contrast, fine-tuning's Precision varies more widely (0.41 on Finnish to 0.54 on Estonian), with higher F1-scores on Polish (0.47) and Estonian (0.65) but a noticeable drop on Finnish (0.45), suggesting a tendency to over-adapt to the training domain. This domain-specific limitation could reduce its effectiveness in regions with differing spectral or structural characteristics, whereas \textit{Feature-level KD}'s stable performance ensures reliability across varied ecosystems. t-SNE visualizations and cluster metrics reinforce this advantage, showing reduced domain separation (e.g., higher overlap ratio of 0.491 vs. 0.220 for \textit{Baseline} pairs on Polish) and greater compactness, indicating more mixed and invariant representations that support its cross-domain applicability.

\subsection{Data Efficiency: Performance in Low-Data Scenarios}
In low-data settings, \textit{Fine-tuned} retains a higher F1-score (0.59 at 25\% data) compared to \textit{Feature-level KD} (0.56), with stronger Recall (0.73 vs. 0.66), showcasing its adaptability to resource-constrained environments where maximizing detection is key, such as initial surveys in under-resourced regions. However, \textit{Feature-level KD} sustains competitive Precision (0.48 vs. 0.50 at 25\%) and significantly reduces false positives (1299 vs. 1658), offering a practical advantage in scenarios where precision and error reduction are critical despite scarce labeled imagery. This balance makes \textit{Feature-level KD} a strong contender for remote or under-resourced forest regions, where minimizing false positives can optimize limited field efforts. Linear probing further highlights KD's efficiency, revealing superior separability in low-data adaptations (e.g., higher Layer 1 accuracy of 0.81 vs. 0.78 on Polish), which correlates with its representational stability and supports its effectiveness across varying data conditions.

\subsection{Practical Implications for Forestry}
For forestry management, \textit{Feature-level KD}'s high Precision and spatial accuracy align exceptionally well with tasks requiring reliable, targeted detection of dead trees, such as pinpointing individual trees for selective logging, pest outbreak containment, or fire risk mitigation. Its ability to minimize false positives reduces operational costs and enhances efficiency, enabling more focused interventions that conserve resources and improve decision-making in ecological monitoring. In contrast, fine-tuning, while better suited for exhaustive detection across large areas, may be less practical due to its higher error rate, which could lead to misdirected efforts and increased resource expenditure. \textit{Feature-level KD} thus strikes a balance between comprehensive detection and precision, offering a scalable solution for cross-domain forest health monitoring that addresses both immediate needs and long-term sustainability goals.

\subsection{Potential for On-Board Deployment}
In traditional knowledge distillation, the student model is typically more compact than the teacher model, making it an ideal candidate for resource-constrained platforms such as drones or edge devices used in on-board processing~\citep{himeur2024applications}. However, in this study, the student and teacher models share the same architecture, meaning the student model does not inherently offer reductions in model size or computational complexity. Despite this limitation, the distillation process significantly enhances the student model's generalization across diverse domains, as demonstrated by its consistent performance on datasets from Finnish, Polish, German, and Estonian forests. This robustness is particularly critical for on-board deployment, where systems must reliably detect dead trees in varied, potentially unseen environments under real-time constraints. Future work could explore integrating KD with compression techniques such as pruning or quantization to further optimize the student model for resource-limited settings, building on its already strong linear separability in probing analyses and potentially enabling deployment on lightweight hardware for widespread forest monitoring.

\subsection{Limitations and Future Directions}
This study is limited to four different datasets encapsulating boreal and temperate forest types, which may not fully represent other ecosystems such as tropical or arid regions. While the model performs well within these environments, its effectiveness in more diverse settings remains untested, potentially limiting its global applicability. Future work could explore expanding the dataset to include a broader range of forest types and conditions, incorporating advanced imaging techniques to enhance the model's generalizability across global forest ecosystems. Additionally, the limited exploitation of multi-spectral features (currently restricted to RGB-NIR) and potential annotation biases from expert labeling warrant further investigation. Incorporating hyperspectral data could reveal subtler spectral signatures of decay, while assessing inter-annotator agreement could improve annotation reliability, paving the way for more robust and versatile dead tree detection systems.

While fine-tuning excels in raw detection performance, \textit{Feature-level KD} emerges as the most effective KD variant, offering superior precision, spatial accuracy, and cross-domain consistency. Its advantages make it particularly suited for precision-critical forestry applications, advancing the field of ecological monitoring with an efficient, scalable approach to dead tree detection.

\section{Conclusion}
\label{sec:conclusion}

This study underscores the efficacy of KD techniques in optimizing the \oursone model for dead tree detection across varied forest domains, with \textit{Feature-level KD} emerging as the most effective variant. While fine-tuning delivers competitive detection performance on the target domain with an Instance F1-score of 0.58 and Instance Recall of 0.80, its elevated false positive rate (1765 FPs) and variable cross-domain consistency limit its suitability for resource-constrained forestry applications. In contrast, \textit{Feature-level KD} achieves a superior balance of Instance Precision (0.55), spatial accuracy (Mean Centroid Error: 3.039), and cross-domain robustness, with significantly fewer false positives (1134 FPs) and competitive true positives (1384 TP). Representational analyses further highlight its invariance: higher deep-layer alignments (e.g., CKA 0.488 in Layer 4 vs. fine-tuning's 0.419), enhanced domain mixing (t-SNE overlap 0.491), and stronger linear separability (probing AUC 0.95 in Layer 4), explaining its consistent precision (e.g., 0.60 on German) and data efficiency even in low-data scenarios.

The findings underscore KD's potential to address challenges such as domain variability and limited data in remote sensing, enhancing transfer learning for ecological monitoring. Nevertheless, the study's reliance on boreal and temperate datasets, limited use of multi-spectral features (RGB-NIR only), and potential annotation biases suggest areas for improvement. Future research could explore hybrid KD with unsupervised adaptation, advanced architectures (e.g., lighter students for edge deployment), and broader datasets encompassing diverse ecosystems to boost performance and generalizability. In summary, this work lays a foundation for scalable, efficient solutions in forest health management, advancing the broader mission of sustainable ecological conservation.

\section*{Acknowledgments}

This work was funded by the European Union (ERC-2023-STG grant agreement no. 101116404). Views and opinions expressed are, however, those of the author(s) only and do not necessarily reflect those of the European Union or the European Research Council Executive Agency. Neither the European Union nor the granting authority can be held responsible for them. The authors greatly acknowledge CSC -- IT Center for Science, Finland, for computational resources.

\bibliographystyle{unsrtnat}
\bibliography{references} 

\end{document}